\documentclass[letterpaper, 10 pt, conference]{ieeeconf}  

\IEEEoverridecommandlockouts                              

\usepackage[utf8]{inputenc} 
\usepackage[T1]{fontenc}    
\usepackage{hyperref}       
\usepackage{url}            
\usepackage{booktabs}       
\usepackage{amsfonts}       
\usepackage{nicefrac}       
\usepackage{microtype}      
\usepackage{gensymb}

\usepackage{comment}
\usepackage{epsfig}
\usepackage{epstopdf}
\usepackage{graphics}
\usepackage{subfigure}

\usepackage{times}
\usepackage[table]{xcolor}
\usepackage{soul}
\usepackage[utf8]{inputenc}
\usepackage{caption}

\usepackage{amsmath}

\definecolor{lightgray}{gray}{0.85}
\DeclareMathOperator*{\argmin}{arg\,min}

\newif\ifcommenton
\commentontrue
\ifcommenton
\newcommand{\red}[1]{\textcolor{red}{#1} }
\else
\newcommand{\red}[1]{}
\fi

\title{\LARGE \bf
Multimodal Trajectory Predictions for Autonomous \\ Driving using Deep Convolutional Networks
}

\author{Henggang Cui, Vladan Radosavljevic, Fang-Chieh Chou, Tsung-Han Lin, \\ Thi Nguyen, Tzu-Kuo Huang, Jeff Schneider and Nemanja Djuric$^{1}$
\thanks{$^{1}$Authors are with Uber Advanced Technology Group,
       50 33rd Street, Pittsburgh, USA; corresponding author email
        {\tt\small ndjuric@uber.com}}
}

\begin{document}

\maketitle
\thispagestyle{empty}
\pagestyle{empty}

\begin{abstract}
Autonomous driving presents one of the largest problems that the robotics and artificial intelligence communities are facing at the moment, both in terms of difficulty and potential societal impact. Self-driving vehicles (SDVs) are expected to prevent road accidents and save millions of lives while improving the livelihood and life quality of many more. However, despite large interest and a number of industry players working in the autonomous domain, there still remains more to be done in order to develop a system capable of operating at a level comparable to best human drivers. One reason for this is high uncertainty of traffic behavior and large number of situations that an SDV may encounter on the roads, making it very difficult to create a fully generalizable system. To ensure safe and efficient operations, an autonomous vehicle is required to account for this uncertainty and to anticipate a multitude of possible behaviors of traffic actors in its surrounding. We address this critical problem and present a method to predict multiple possible trajectories of actors while also estimating their probabilities. The method encodes each actor's surrounding context into a raster image, used as input by deep convolutional networks to automatically derive relevant features for the task. Following extensive offline evaluation and comparison to state-of-the-art baselines, the method was successfully tested on SDVs in closed-course tests.

\end{abstract}

\section{Introduction}
Recent years have witnessed unprecedented progress in Artificial Intelligence (AI) applications, with smart algorithms rapidly becoming an integral part of our daily lives. The AI methods are used by hospitals to help diagnose diseases \cite{amato2013artificial}, matchmaking services are using learned models to connect potential couples \cite{albury2017data}, and social media feeds are built by algorithmic approaches \cite{katakis2009adaptive}, to name just a few affecting millions of people. Nevertheless, despite huge strides the AI revolution is far from over, and is likely to further accelerate in the coming years \cite{harari2017reboot}. Interestingly, one of the major industries mostly undisturbed by the ongoing progress is the automobile domain, where thus far AI has seen limited use. Large car-makers made some advances by using AI within Advanced Driver-Assistance Systems (ADAS) \cite{lindgren2006state}, however, its full power remains to be harnessed through the advent of new smart technologies, such as self-driving vehicles (SDVs).

Driving a vehicle in traffic is a surprisingly dangerous task considering it is such a common activity of many, even for human drivers with several years of experience \cite{curry2015young}. While the car manufacturers are working hard on improving vehicle safety through better design and ADAS systems, grim year-to-year statistics indicate that there is still a lot more to be done in order to revert the negative trends observed on public roads. In particular, car accidents amounted to more than $5\%$ of deaths in the US in 2015 \cite{national2017health}, with the human factor to blame in the vast majority of the crashes \cite{singh2015critical}. This is unfortunately not a recent problem, and researchers have been trying to understand the reasons and causes for several decades. Studies include investigating effect of driver distractions \cite{wilson2010trends}, alcohol and drug use \cite{barbone1998association,borkenstein1974role}, and driver's age \cite{curry2015young} among other factors, as well as how to most effectively impact drivers' behavior accepting that they are fallible \cite{naatanen1976road}. Not surprisingly, a common theme in the existing body of literature is that humans are the most unreliable part of the traffic system, which could be mitigated through the development and wide adoption of SDVs. This prospect is made possible by the latest breakthroughs in hardware and software technologies, opening doors for the fields of robotics and AI to potentially make their greatest societal impact yet.

Self-driving technology has been under development for a long time, with the earliest attempts going as far back as the 1980s and the work on ALVINN \cite{pomerleau1989alvinn}. However, only recently the technological advances reached a point where wider use could be possible, as exemplified by the results of 2007 DARPA Urban Challenge \cite{montemerlo2008junior,urmson2008self}. Here, the participating teams were required to navigate complex urban surroundings and handle conditions commonly encountered on public roads, as well as interact with human- and robot-driven vehicles.
These early successes spurred large interest in the autonomous domain that we see today, and a number of industry players (such as Uber or Waymo) and governmental bodies are racing to set up technological and legal scaffolding for SDVs to become a reality. Nevertheless, despite the progress achieved thus far, there remains more to be done to make SDVs operate at the human level and fully commercialize them.

To safely and effectively operate in the real world, a critical piece of the autonomous puzzle is to correctly predict movement of surrounding actors,
and a successful system also needs to account for their inherent multimodal nature.
We focus on this task and build upon our deployed deep learning-based work \cite{dp2018}, which creates bird's-eye view (BEV) rasters encoding high-definition map and surroundings to predict actor's future, and present the following contributions:
\begin{itemize}
  \item we extend the state-of-the-art and propose a method that goes beyond inferring a single trajectory, and instead reasons about multiple trajectories and their probabilities;
  \item following extensive offline study of multi-hypothesis methods, the proposed method was successfully tested onboard SDVs in closed-course tests.
\end{itemize}

Illustrative examples of how multimodality of future $6$-second trajectories is captured by our model are shown in Figure \ref{fig:example_output}. The method uses rasterized vehicle context (including the high-definition map and other actors) as a model input to predict actor's movement in a dynamic environment \cite{dp2018}. As the vehicle is approaching the intersection the multimodal model (where we set the number of modes to $2$) estimates that going straight has slightly less probability than a right turn, see Figure \ref{fig:example_output}a. After three timesteps the vehicle continued to move straight, at which point probability of right turn drops significantly (Figure \ref{fig:example_output}c); note that in actuality the vehicle continued going straight through the intersection. We can see that a single-modal model is not capable of capturing multimodality of the scene, and instead roughly predicts mean of the two modes, as illustrated in Figures \ref{fig:example_output}b and \ref{fig:example_output}d.

\section{Related work}

%
%
%
%
%
%


The problem of predicting actor's future motion has been addressed in a number of recent publications. Comprehensive overview of the topic can be found in \cite{Lefevre2014, Wiest_2017}, and in this section we review relevant work from the perspective of autonomous driving. First, we cover engineered approaches applied in practice in the self-driving industry. Then, we discuss machine learning approaches for movement prediction, with a particular emphasis on deep learning methods.

\subsection{Actor motion prediction in self-driving systems}


Most of the deployed self-driving systems use well-established engineered approaches for actors' motion prediction. The common approach consists of computing object's future motion by propagating its state over time based on assumptions of underlying physical system and using techniques such as Kalman filter (KF) \cite{Kalman1960, CosgunMCHDALTA17}. While this approach works well for short-term predictions, its performance degrades for longer horizons as the model ignores surrounding context (e.g., roads, other actors, traffic rules). Addressing this issue, method proposed by Mercedes-Benz \cite{Bertha2015} uses map information as a constraint to compute vehicle's future position at longer term. The system first associates each detected vehicle with one or more lanes from the map. Then, all possible paths are generated for each {\it (vehicle, associated lane)} pair based on map topology, lane connectivity, and vehicle's current state estimate. This heuristic provides reasonable predictions in common cases, however it is sensitive to errors in association of vehicles and lanes. As an alternative to existing deployed engineered approaches, the proposed method automatically learns from the data that vehicles typically obey road and lane constraints, while generalizing well to various situations observed on the roads. In addition, we propose an extension of our method that incorporates existing lane-association ideas.


\begin{figure}[t!]
\centering
\subfigure[Two-modal, time $T$]{
\includegraphics[keepaspectratio=1,height=0.42\columnwidth]{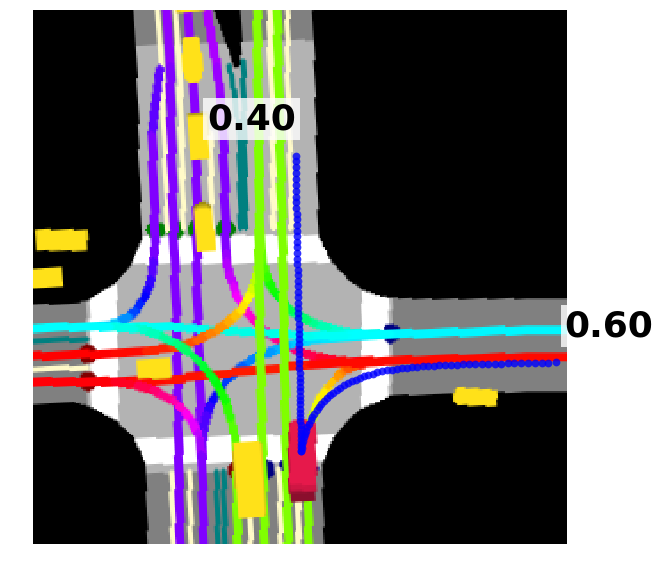}
}
\subfigure[Single-modal, time $T$]{
\includegraphics[keepaspectratio=1,height=0.42\columnwidth]{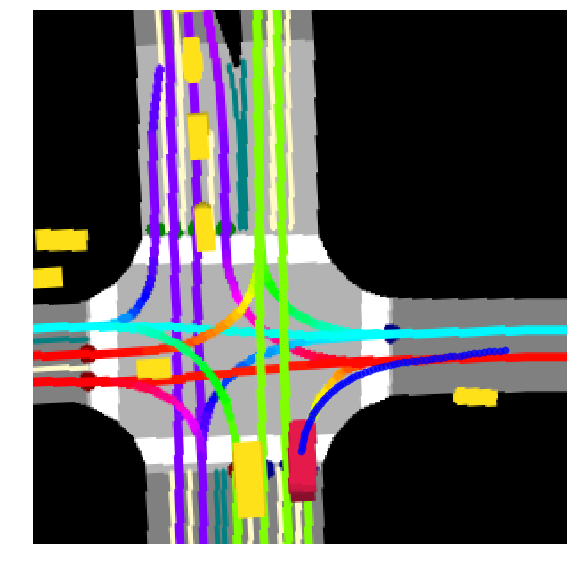}
}
\subfigure[Two-modal, time $T+3$]{
\includegraphics[keepaspectratio=1,height=0.42\columnwidth]{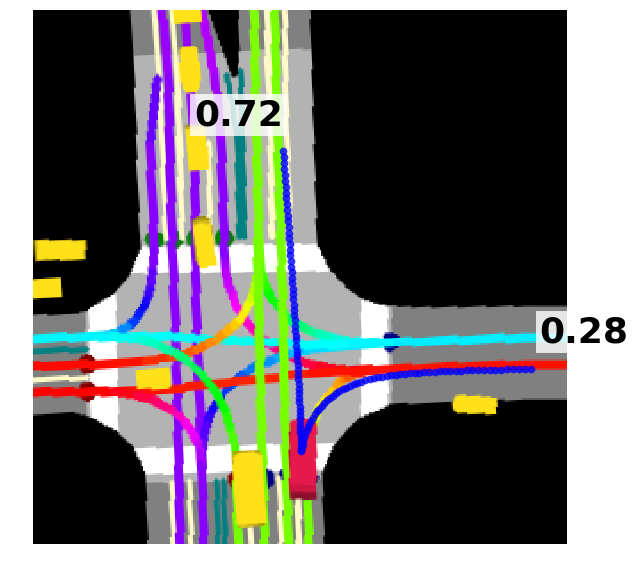}
}
\subfigure[Single-modal, time $T+3$]{
\includegraphics[keepaspectratio=1,height=0.42\columnwidth]{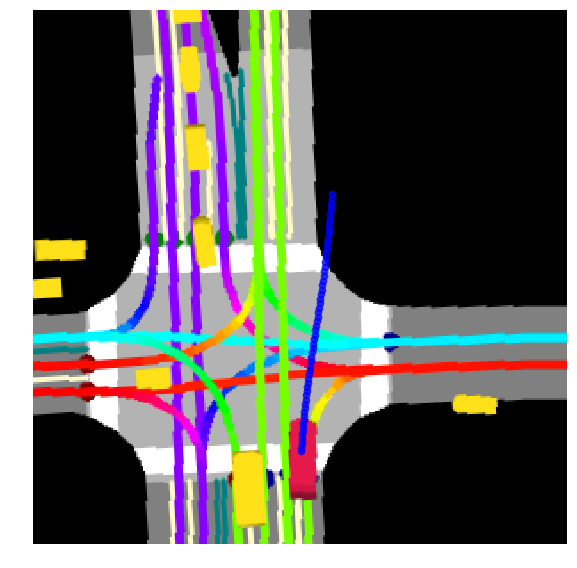}
}
\caption{Modeling multimodality of future $6$-second vehicle trajectories; predicted trajectories are marked in blue, with their probabilities indicated at the end of the trajectories}
\label{fig:example_output}
\end{figure}

\subsection{Machine-learned prediction models}
Manually engineered models fail to scale to many different traffic scenarios, which motivated the use of machine learning models as alternatives, such as Hidden Markov Model \cite{Streubel2014}, Bayesian networks \cite{Schreier2016}, or Gaussian Processes \cite{Wang2008}. In recent work researchers focused on how to model environmental context using Inverse Reinforcement Learning (IRL) \cite{ng2000algorithms}. Kitani {\it et al.} \cite{Kitani2012} used inverse optimal control to predict pedestrian paths by considering scene semantics, however the existing IRL methods are inefficient for real-time applications. 

\begin{figure*}[!t]
\centering
\includegraphics[keepaspectratio=1,width=1.4\columnwidth]{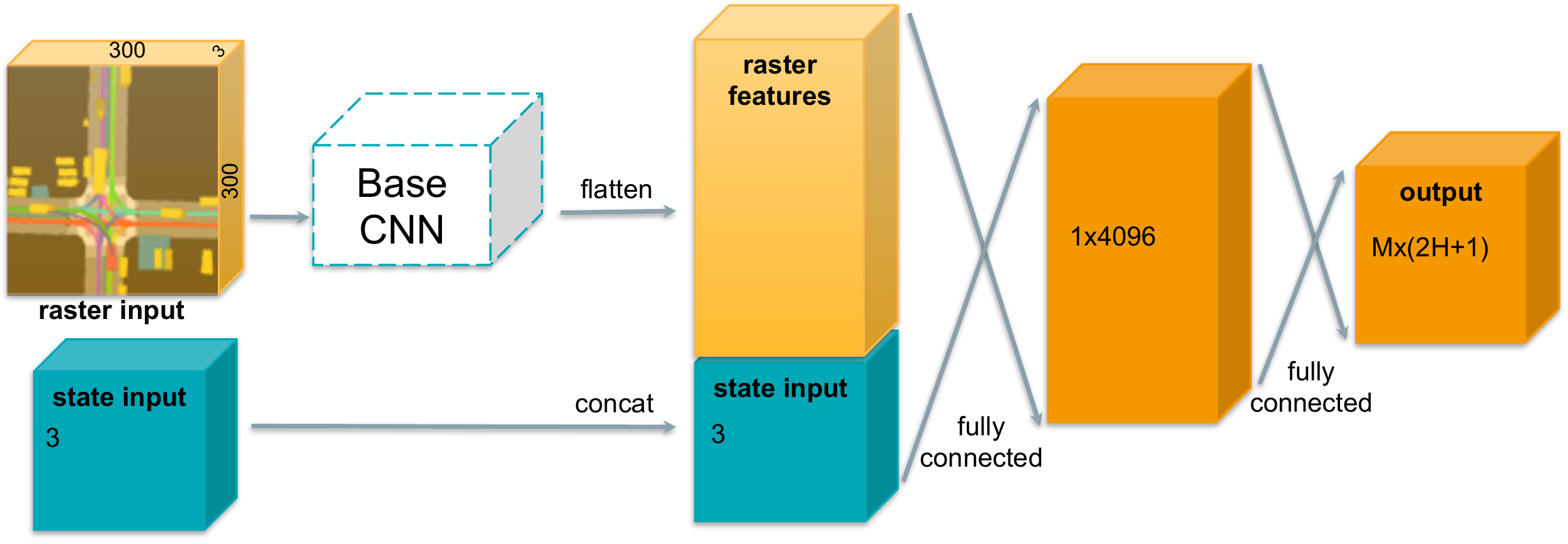}
\caption{The proposed network architecture}
\label{fig:network_diag}
\vspace{-.15in}
\end{figure*}

The success of deep learning in many real-life applications \cite{hinton2015} prompted research on its use for motion prediction. One line of research follows the recent success of recurrent neural networks (RNN), namely Long Short-Term Memory (LSTM), for sequence prediction tasks. Authors of \cite{Alahi2016, Gupta_2018_CVPR} applied LSTM to predict pedestrians' future trajectories with social interactions. In \cite{kim2017probabilistic}, LSTM was applied to predict vehicle locations using past trajectory data. In \cite{Lee2017}, another RNN-variant called gated recurrent unit (GRU) combined with conditional variational auto-encoder (CVAE) was used to predict vehicle trajectory. Alternatively, \cite{Fragkiadaki2016, Watters_2017} predicted motion of simple physical systems directly from image pixels by applying Convolutional NN (CNN) to a sequence of visual glimpses. In \cite{dp2018} authors presented a system where short-term vehicle trajectories were predicted using CNNs, taking as input a BEV raster image encoding individual actor's surrounding context, which was later applied to vulnerable road users as well \cite{chou2018mlits}. Despite their success these methods do not address potential multimodality of possible future trajectories, required for accurate long-term traffic predictions.

There exists a number of studies addressing the problem of modeling multimodality. Mixture Density Networks (MDNs) \cite{bishop1994mixture} are conventional neural networks which solve multimodal regression tasks by learning parameters of a Gaussian mixture model. However, MDNs are often difficult to train in practice due to numerical instabilities when operating in high-dimensional spaces. To overcome this problem researchers proposed training an ensemble of networks \cite{NIPS2016_6270}, or a single network to produce $M$ different outputs for $M$ different hypotheses \cite{rupprecht2017learning} using a loss that only accounts for the closest prediction to ground truth labels. Due to good empirical results our work builds upon these efforts. Furthermore, in \cite{deo2018multi} the authors introduced a method for multimodal trajectory prediction for highway vehicles by learning a model that assigns probabilities to six maneuver classes. The approach requires a predefined discrete set of possible maneuvers, which may be hard to define for complex city driving. Alternatively, in \cite{Alahi2016, Gupta_2018_CVPR, Lee2017} the authors proposed to generate multimodal predictions through sampling, which requires repeated forward passes to generate multiple trajectories. On the other hand, our proposed approach computes multimodal predictions directly, in a single forward-pass of a CNN model.


\section{Proposed approach}
In this section we discuss the proposed method for predicting multimodal trajectories of traffic actors. We introduce the problem setting and notation, followed by the discussion of our CNN architecture and the considered loss functions.

\subsection{Problem setting}
Let us assume that we have access to real-time data streams coming from sensors such as lidar, radar, or camera, installed aboard a self-driving vehicle. In addition, assume that these inputs are used by an existing detection and tracking system, outputting state estimates $\mathcal{S}$ for all surrounding actors (state comprises the bounding box, position, velocity, acceleration, heading, and heading change rate). Denote a set of discrete times at which tracker outputs state estimates as $\mathcal{T} = \{t_1, t_2, \dots, t_T\}$, where time gap between consecutive time steps is constant (e.g., the gap is equal to $0.1s$ for tracker running at a frequency of $10Hz$). Then, we denote state output of a tracker for the $i$-th actor at time $t_j$ as ${\bf s}_{ij}$, where $i = 1, \dots, N_j$ with $N_j$ being a number of unique actors tracked at $t_j$. Note that in general the actor counts vary for different time steps as new actors appear within and existing ones disappear from the sensor range. 
Moreover, we assume access to detailed, high-definition map information $\mathcal{M}$ of the SDV's operating area, including road and crosswalk locations, lane directions, and other relevant map information.

\subsection{Modeling multimodal trajectories}
Following our previous work~\cite{dp2018},
we first rasterize an actor-specific BEV raster image encoding the actor's map surrounding and neighboring actors (e.g., other vehicles and pedestrians),
as exemplified in Figure \ref{fig:example_output}.
Then, given $i$-th actor's raster image and state estimate ${\bf s}_{ij}$ at time step $t_j$,
we use a CNN model to predict a multitude of $M$ possible future state sequences $\{[{\tilde {\bf s}}_{im(j+1)}, \dots, {\tilde {\bf s}}_{im(j+H)}]\}_{m=1, \ldots, M}$, as well as each sequence's probability $p_{im}$ such that $\sum_m p_{im} = 1$, where $m$ indicates mode index and $H$ denotes the number of future consecutive time steps for which we predict states (or {\it prediction horizon}).
For a detailed description of the rasterization method, we refer the reader to our previous work \cite{dp2018}.
Without the loss of generality, in this work we simplify the task to infer $i$-th actor's future $x$- and $y$-positions instead of full state estimates, while the remaining states can be derived by considering ${\bf s}_{ij}$ and the future position estimates. Both past and future positions at time $t_j$ are represented in the actor-centric coordinate system derived from actor's state at time $t_j$, where forward direction is $x$-axis, left-hand direction is $y$-axis, and actor's bounding box centroid is the origin.

The proposed network is illustrated in Figure \ref{fig:network_diag}. It takes an actor-specific $300 \times 300$ RGB raster image with $0.2m$ resolution and actor's current state (velocity, acceleration, and heading change rate) as input, and outputs $M$ modes of future $x$- and $y$-positions ($2H$ outputs per mode) along with their probabilities (one scalar per mode). This results in $(2H+1)M$ outputs per actor. Probability outputs are passed through a softmax layer to ensure they sum to $1$. Note that any CNN architecture can be used as the base network; following results in \cite{dp2018} we use MobileNet-v2~\cite{sandler2018inverted}.

\begin{figure}[!t]
\centering
\includegraphics[keepaspectratio=1,width=0.6\columnwidth]{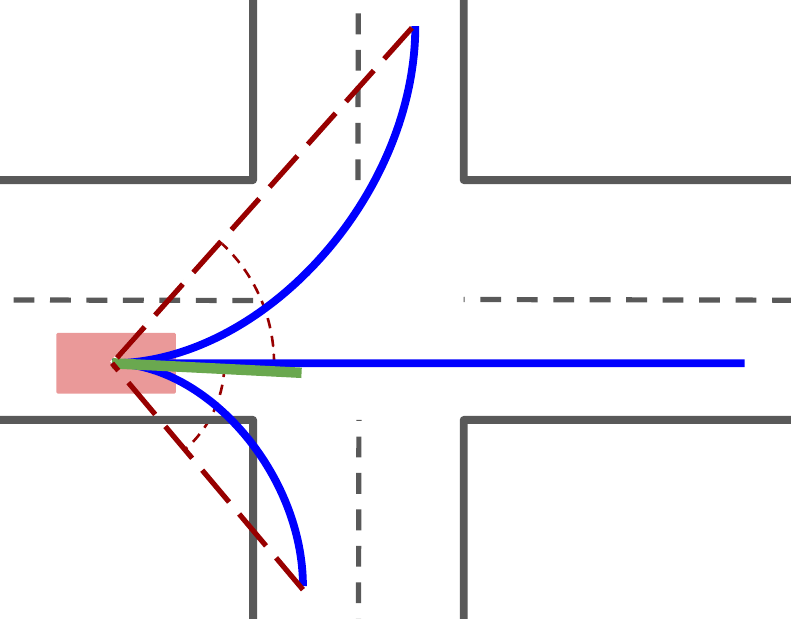}
\caption{Mode selection methods (modes shown in blue): the ground truth (green) matches to the right-turn mode when using displacement, but the straight mode when using angle}
\label{fig:disp_vs_angle}
\vspace{-.1in}
\end{figure}

\subsection{Multimodal optimization functions}
In this section we discuss loss functions that we proposed to model the inherent multimodality of the trajectory prediction problem.
First, let us define a single-mode loss $L$ of the $i$-th actor's $m$-th mode at time $t_j$ as average displacement error (or $\ell_2$-norm) between the points of ground-truth trajectory ${\boldsymbol \tau}_{ij}$ and predicted trajectory of the $m$-th mode ${\tilde {\boldsymbol \tau}}_{imj}$,
\begin{equation}
\label{eq:unimodal_loss}
L({\boldsymbol \tau}_{ij}, {\tilde {\boldsymbol \tau}}_{imj}) = \frac{1}{H} \sum_{h=1}^{H} \|\tau_{ij}^h - {\tilde \tau}_{imj}^h\|_2,
\end{equation}
where $\tau_{ij}^h$ and ${\tilde \tau}_{imj}^h$ are $2$-D vectors representing $x$- and $y$-positions at horizon $h$ of ${\boldsymbol \tau}_{ij}$ and ${\tilde {\boldsymbol \tau}}_{imj}$, respectively.

One straightforward multimodal loss function we can use is the \emph{Mixture-of-Experts} (ME) loss, defined as
\begin{equation}
\label{eq:me_loss}
\mathcal{L}^{\mathrm{ME}}_{ij} = \sum_{m=1}^M p_{im} L({\boldsymbol \tau}_{ij}, {\tilde {\boldsymbol \tau}}_{imj}),
\end{equation}
corresponding to the expected displacement loss.
However, as shown by our evaluation results in Section~\ref{sect:exp}, the ME loss is not suitable for the trajectory prediction problem due to the mode collapse problem.
To address this issue, we propose to use a novel \emph{Multiple-Trajectory Prediction} (MTP) loss,
motivated by~\cite{rupprecht2017learning},
that explicitly models the multimodality of the trajectory space.
In the MTP method, for the $i$-th actor at time $t_j$ we first run the forward pass of the neural network to obtain $M$ output trajectories. We then identify mode $m^*$ that is closest to the ground-truth trajectory according to an arbitrary trajectory distance function $dist({\boldsymbol \tau}_{ij}, {\tilde {\boldsymbol \tau}}_{imj})$,
\begin{equation}
\label{eq:best_m}
m^* = \argmin_{m \in \{1, \ldots, M\}} dist({\boldsymbol \tau}_{ij}, {\tilde {\boldsymbol \tau}}_{imj}).
\end{equation}
After selecting the best matching mode $m^*$, the final loss function can be defined as follows,
\begin{equation}
\label{eq:MTP_loss}
\mathcal{L}^{\mathrm{MTP}}_{ij} = \mathcal{L}^{class}_{ij} + \alpha \sum_{m=1}^M I_{m = m^*} L({\boldsymbol \tau}_{ij}, {\tilde {\boldsymbol \tau}}_{imj}),
\end{equation}
where $I_{c}$ is a binary indicator function equal to $1$ if the condition $c$ is true and 0 otherwise, $\mathcal{L}^{class}_{ij}$ is a classification cross-entropy loss defined as follows,
\begin{equation}
\label{eq:class_loss}
\mathcal{L}^{class}_{ij} = -\sum_{m=1}^M I_{m = m^*} \log p_{im},
\end{equation}
and $\alpha$ is a hyper-parameter used to trade-off the two losses. In other words, we force the probability of the best matching mode $m^*$ to be as close as possible to $1$, and push probabilities of the other modes to $0$. Note that during training the position outputs are updated only for the winning mode, while the probability outputs are updated for all the modes.
This causes each mode to specialize for a distinct class of actor behavior (e.g., going straight or turning), and successfully addresses the mode collapse problem as shown later in the experiments.

\begin{figure}[!t]
\centering
\includegraphics[keepaspectratio=1,width=1.0\columnwidth,trim={0 0 0 1.0cm},clip]{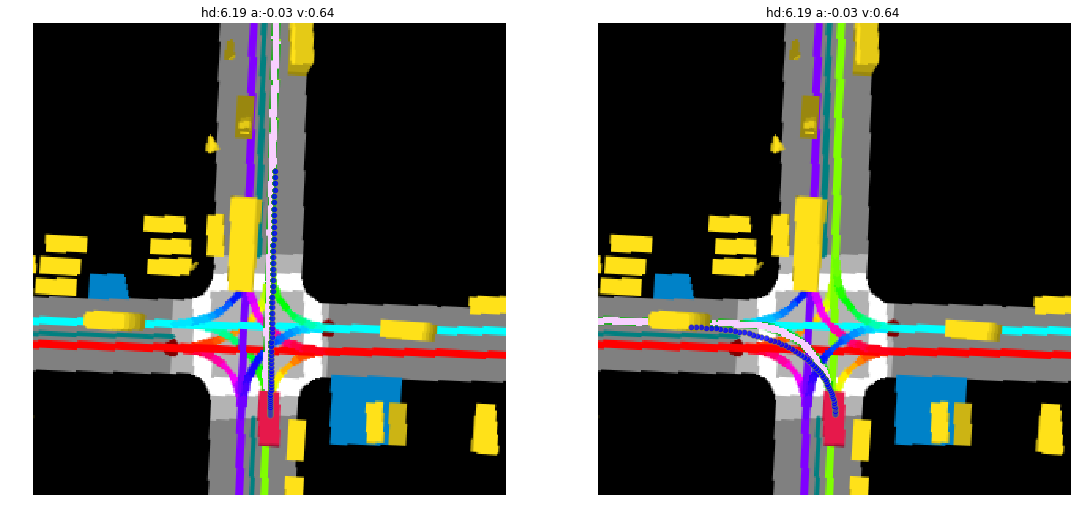}
\caption{Example of trajectories output by LF model in a same scene for different following lanes (represented in light pink)}
\label{fig:goal_dp_example}
\vspace{-.1in}
\end{figure}

We experimented with several different trajectory distance functions $dist({\boldsymbol \tau}_{ij}, {\tilde {\boldsymbol \tau}}_{imj})$.
In particular, for the first option we used the average displacement \eqref{eq:unimodal_loss} between the two trajectories.
However, this distance function does not model the multimodal behaviors well at intersections, as illustrated in Figure \ref{fig:disp_vs_angle}.
To address this problem, we propose a function that measures distance by considering an angle between the last points of the two trajectories as seen from the actor position, which improves handling of the intersection scenarios\footnote{In practice, we put a threshold of $5\degree$ on the angle difference. Then, all modes which differ from the ground-truth trajectory by less than $5\degree$ are considered to be potential matches, and we use displacement distance to break the tie among them. We found that this simple optimization makes the model generalize well for both the intersection and straight-road scenarios.}.
The quantitative comparison results are given in Section~\ref{sect:exp}.

Lastly, for both losses \eqref{eq:me_loss} and \eqref{eq:MTP_loss} we train the CNN parameters $\theta$ to minimize loss over the training data,
\begin{equation}
\label{eq:training_loss}
\theta^* = \argmin_{\theta} \sum_{j=1}^T \sum_{i=1}^{N_j} \mathcal{L}_{ij}^{\bullet}.
\end{equation}

Note that that our multimodal loss functions are agnostic to the choice of the per-mode loss function $L({\boldsymbol \tau}_{ij}, {\tilde {\boldsymbol \tau}}_{imj})$, and it is straightforward to extend our method to predict trajectory point uncertainties by using the negative Gaussian log-likelihood, as proposed in \cite{dp2018}.

\begin{table*} [!ht]
\caption{Comparison of prediction errors for competing methods (in meters)}
\label{tab:pred-errors}
\centering
{
  \begin{tabular}{ccccccccccc}
     & & \multicolumn{3}{c}{\bf Displacement [m]} & \multicolumn{3}{c}{\bf Along-track [m]} & \multicolumn{3}{c}{\bf Cross-track [m]} \\
    {\bf Method} & {\bf No. modes} & {\bf @1s} & {\bf @6s} & {\bf Average} & {\bf @1s} & {\bf @6s} & {\bf Average} & {\bf @1s} & {\bf @6s} & {\bf Average} \\
    \hline
    \rowcolor{lightgray}
    UKF & 1 & 0.54 & 10.58 & 3.99 & 0.41 & 7.88 & 2.94 & 0.26 & 5.05 & 1.94 \\
    \rowcolor{lightgray}
    STP & 1 & 0.34 & 4.14 & 1.54 & 0.30 & 3.91 & 1.45 & 0.11 & 0.72 & 0.30 \\
    \hline
    ME & 2 & 0.34 & 4.17 & 1.55 & 0.30 & 3.93 & 1.45 & 0.11 & 0.74 & 0.30 \\
    ME & 3 & 0.34 & 4.13 & 1.54 & 0.30 & 3.90 & 1.44 & 0.11 & 0.72 & 0.30 \\
    ME & 4 & 0.34 & 4.13 & 1.54 & 0.30 & 3.89 & 1.44 & 0.11 & 0.73 & 0.30 \\
    \hline
    \rowcolor{lightgray}
    MDN & 2 & 0.37 & 4.18 & 1.58 & 0.33 & 3.86 & 1.46 & 0.10 & 0.61 & 0.27 \\
    \rowcolor{lightgray}
    MDN & 3 & 0.27 & 3.31 & 1.26 & 0.22 & 2.95 & 1.12 & {\bf 0.09} & 0.60 & 0.26 \\
    \rowcolor{lightgray}
    MDN & 4 & 0.27 & 2.91 & 1.18 & 0.21 & 2.55 & 1.05 & {\bf 0.09} & 0.59 & 0.26 \\
    \hline
    MTP w/ disp. & 2 & 0.25 & 2.72 & 1.07 & 0.20 & 2.49 & 0.97 & {\bf 0.09} & 0.49 & {\bf 0.23} \\
    MTP w/ disp. & 3 & {\bf 0.23} & {\bf 2.31} & {\bf 0.94} & {\bf 0.17} & {\bf 2.05} & {\bf 0.82} & {\bf 0.09} & {\bf 0.46} & {\bf 0.23} \\
    MTP w/ disp. & 4 & 0.24 & 2.51 & 1.00 & 0.18 & 2.23 & 0.88 & {\bf 0.09} & 0.51 & 0.24 \\
    \hline
    \rowcolor{lightgray}
    MTP w/ angle & 2 & 0.26 & 2.80 & 1.10 & 0.21 & 2.56 & 1.00 & {\bf 0.09} & 0.51 & 0.24 \\
    \rowcolor{lightgray}
    MTP w/ angle & 3 & {\bf 0.23} & 2.33 & 0.95 & {\bf 0.17} & 2.08 & 0.84 & {\bf 0.09} & {\bf 0.46} & {\bf 0.23} \\
    \rowcolor{lightgray}
    MTP w/ angle & 4 & 0.25 & 2.57 & 1.03 & 0.19 & 2.29 & 0.91 & {\bf 0.09} & 0.51 & 0.24 \\
    \hline
    \hline
\end{tabular}
}
\end{table*}

\subsection{Lane-following multimodal predictions}
Previously we described an approach that explicitly predicts multiple modes in a single forward pass. Following approach in \cite{Bertha2015} where each vehicle is associated to a lane (referred to as a {\it lane-following vehicle}), we propose a method that implicitly outputs multiple trajectories. In particular, assuming a knowledge of the possible lanes that can be followed and a lane-scoring system that filters unlikely lanes, we add another rasterization layer that encodes this information and train the network to output a lane-following trajectory. Then, for one scene we can generate multiple rasters with various lanes to be followed, effectively inferring multiple trajectories. 
To generate training data we first identify lanes that a vehicle actually followed and use that information to create input rasters. We then train the Lane-Following (LF) model using the losses introduced previously by setting $M=1$ (ME and MTP are equal in that case). In practice, LF and other approaches can be used together, separately processing lane-following and remaining actors, respectively. Note that we introduce this approach solely for completeness, as practitioners may find it useful to combine our rasterization ideas with the existing deployed lane-following approaches to obtain multimodal predictions, as described in \cite{Bertha2015}.

In Figure \ref{fig:goal_dp_example} we show rasters for the same scene but using two different following lanes marked in light pink, one going straight and one turning left. The method outputs trajectories that follow the intended paths well, and can be used to generate multitude of trajectories for lane-following vehicles.

\section{Experiments}
\label{sect:exp}

We collected 240 hours of data by manually driving SDV in Pittsburgh, PA and Phoenix, AZ in various traffic conditions (e.g., varying times of day, days of the week). Traffic actors were tracked using Unscented Kalman filter (UKF) \cite{wan2000unscented} with kinematic vehicle model \cite{kong2015kinematic}, taking raw sensor data from the camera, lidar, and radar and outputting state estimates for each tracked vehicle at the rate of $10Hz$. 
The filter is highly optimized and trained on a large amount of labeled data, and has been extensively tested on large-scale, real-world data.
Each actor at each discrete tracking time step amounts to a single data point, with overall data comprising $7.8$ million data points after removing static actors. We considered prediction horizon of $6$ seconds (i.e., $H = 60$), set $\alpha=1$, and used $3$:$1$:$1$ split to obtain train/validation/test data.

We compared the proposed methods to several baselines:
\begin{itemize}
  \item UKF that forward propagates estimated state in time;
  \item Single-trajectory predictor (STP) from \cite{dp2018};
  \item MDN \cite{bishop1994mixture}, as Gaussian mixture over trajectory space,
\begin{equation}
\label{eq:mdn_traj_loss}
\mathcal{L}^{\mathrm{MDN}}_{ij} = -\log \sum_{m=1}^M p_{im} \prod_{h=1}^{H} \mathcal{N}(\tau_{ij}^h | {\tilde \tau}_{imj}^h, {\tilde {\boldsymbol \Sigma}}_{imj}^h),
\end{equation}
where ${\tilde {\boldsymbol \Sigma}}_{imj}^h$ is a $2 \times 2$ covariance matrix measuring uncertainty at horizon $h$, inferred by the network.
\end{itemize}

The models were implemented in TensorFlow \cite{tensorflow2015-whitepaper} and trained on 16 Nvidia Titan X GPU cards. We used open-source distributed framework Horovod \cite{sergeev2018horovod} for training, completing in around 24 hours. We used a per-GPU batch size of $64$ and trained with Adam optimizer \cite{kingma2014adam}, setting initial learning rate to $10^{-4}$ further decreased by a factor of $0.9$ every 20 thousand iterations. All models were trained end-to-end from scratch, and deployed to an SDV with a GPU onboard performing batch inference in about $10ms$ on average.

\begin{figure*}[!t]
\centering
\includegraphics[keepaspectratio=1,width=2.0\columnwidth]{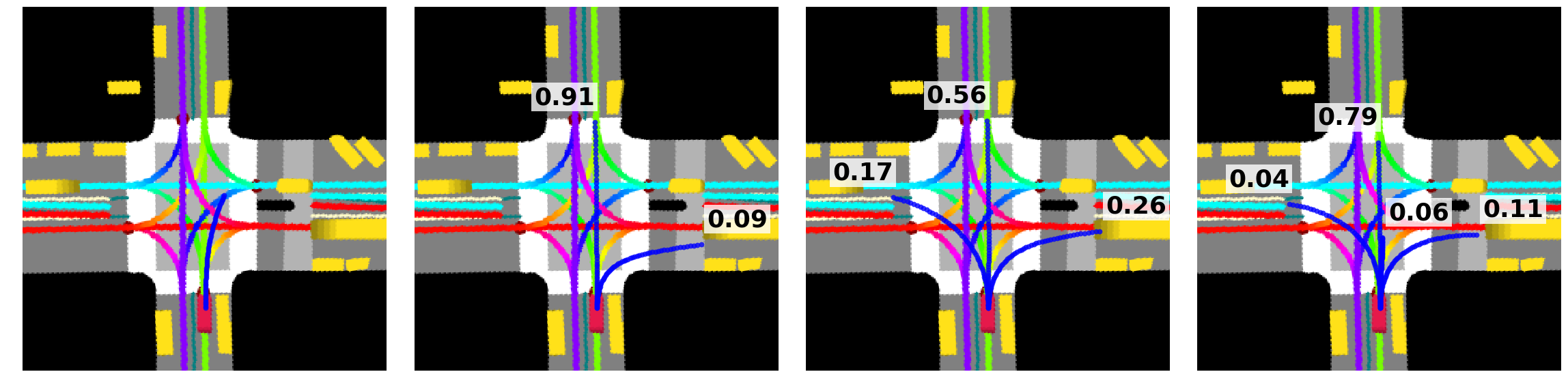}
\vspace{-.1in}
\caption{The effect on output trajectories of varying number of modes $M$ from 1 to 4, respectively from left to right image}
\label{fig:num_modes_effect}
\vspace{-.05in}
\end{figure*}

\subsection{Empirical results}
We compare the considered methods using error metrics relevant for motion prediction: displacement \eqref{eq:unimodal_loss}, as well as along- and cross-track errors \cite{gong2004methodology} measuring longitudinal and lateral deviation from the ground truth, respectively.
As multimodal approaches provide probabilities, one possible evaluation setup is to use prediction error of the most probable mode.
However, earlier work on multimodal predictions~\cite{Lee2017} found that such metric favors single-modal models as they explicitly optimize for the averaged prediction error,
while outputting unrealistic trajectories (see examples in Figure \ref{fig:example_output}).
Mirroring an existing setup from \cite{Lee2017} and \cite{rupprecht2017learning},
we filter out low-probability trajectories (we set the threshold to $0.2$) and
use a lowest-error mode from the remaining set to compute a metric.
We found results computed in this way to be more aligned with an observed performance onboard an SDV.

In Table \ref{tab:pred-errors} we report errors at horizons of $1s$ and $6s$, as well as the averaged metrics over the entire prediction horizon for different numbers of modes $M$ (we vary the mode count from $2$ to $4$).
First, we can see that single-mode models (i.e., UKF and STP) are clearly not suitable for longer-term predictions.
While their short-term predictions at $1$ second are reasonable, $6s$-horizon errors are significantly worse than the best multimodal approaches. This is not unexpected, as in the short term traffic actors are constrained by physics and their immediate surroundings, which results in near-unimodal distribution of the ground truth \cite{dp2018}. On the other hand, at longer horizons multimodality of the prediction problem becomes much more apparent (e.g., as an actor is approaching an intersection there are several discrete choices that they can make given exactly the same scene). Single-modal predictions fail to properly account for that and instead predict a distribution mean, as illustrated in Figure \ref{fig:example_output}.

Furthermore, it is interesting to note that MDN and ME gave similar results as STP for a variety of $M$ values. The reason is a well-known problem of mode collapse,
where only a single mode provides non-degenerate predictions. Thus, in practice an affected multimodal method falls back to being unimodal, failing to fully capture multiple modes. 
Authors of \cite{rupprecht2017learning} reported this issue for MDNs and found multi-hypothesis models to be less affected, as confirmed by our results.

Shifting our focus to the MTP methods, we observe significant improvements over the competing approaches. Both average and $6s$-horizon errors dropped across the board, indicating that the methods are capturing the multimodal nature of the traffic problem. Prediction errors at both short-term $1s$- and long-term $6s$-horizon are lower than for the other methods, although the benefit is greater at longer prediction horizons. Interestingly, the results show that the best performance on all metrics is achieved with $M=3$.

Next, we evaluated different trajectory distance measures for selecting the best matching mode during the MTP training. Table \ref{tab:pred-errors} shows that using displacement as a distance function gives slightly better performance than using angle. However, in order to better understand implications of this choice, we sliced the testing data samples into three categories: turning left, turning right, and going straight ($95\%$ of the test cases are actors going roughly straight, with the remainder evenly split between turns), and report results for $6s$-horizon in Table \ref{tab:pred-errors-slice}.
We can see that using angle improved handling of turns, with a very slight degradation of going-straight cases.
This confirms our hypothesis that the angle matching policy improves the performance at intersections, which is critical to the safety of an autonomous vehicle.
Considering these results, in the remainder of the section we use the MTP model with the angle mode selection policy.

\begin{table} [!t]
\caption{Displacement errors at 6s for different maneuvers }
\label{tab:pred-errors-slice}
\centering
{
  \begin{tabular}{ccccc}
     & & \multicolumn{3}{c}{\bf Displacement @6s [m]} \\
    {\bf Predictor} & {\bf No. modes} & {\bf Left-turn} & {\bf Straight} & {\bf Right-turn} \\
    \hline
    \rowcolor{lightgray}
    MTP w/ disp. & 2 & 4.48 & 2.59 & 5.64 \\
    MTP w/ disp. & 3 & 4.18 & {\bf 2.18} & 5.42 \\
    \rowcolor{lightgray}
    MTP w/ disp. & 4 & 4.51 & 2.37 & 5.47 \\
    \hline
    MTP w/ angle & 2 & 4.65 & 2.67 & 5.66 \\
    \rowcolor{lightgray}
    MTP w/ angle & 3 & {\bf 4.10} & 2.21 & {\bf 5.17} \\
    MTP w/ angle & 4 & 4.91 & 2.43 & 5.52 \\
    \hline
\end{tabular}
}
\vspace{-.10in}
\end{table}

\begin{figure}[!t]
\centering
\includegraphics[keepaspectratio=1,width=0.7\columnwidth]{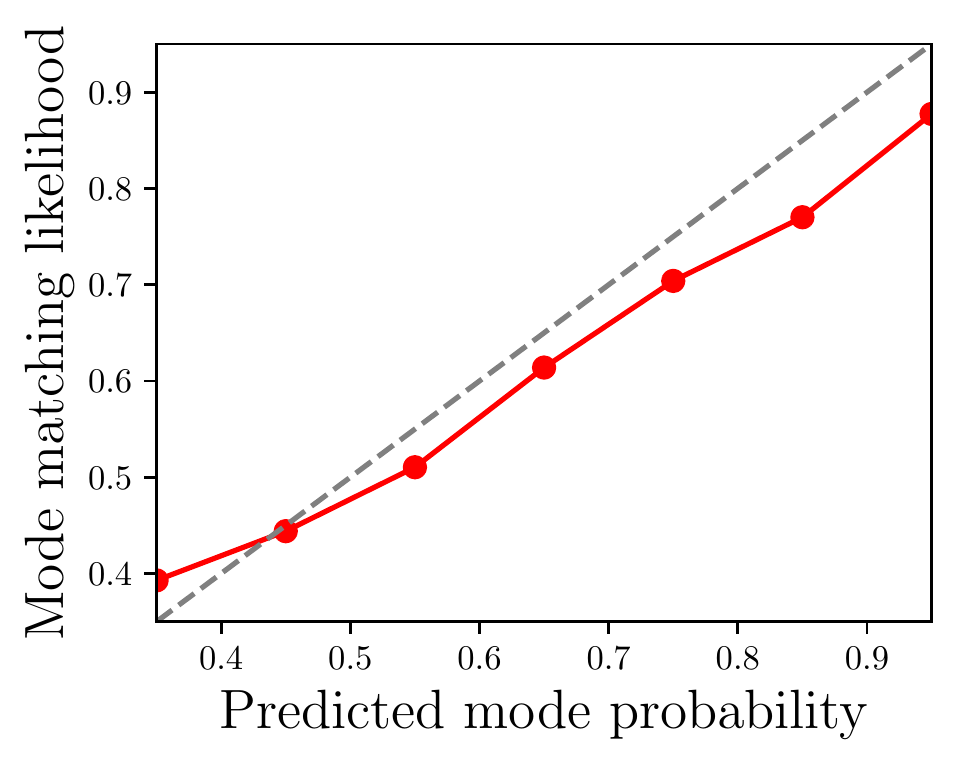}
\caption{Analysis of mode probability calibration}
\label{fig:conf_calib}
\vspace{-.1in}
\end{figure}

In Figure \ref{fig:num_modes_effect} we visualize outputs for an intersection scene with an actor going straight, when increasing number of modes $M$. We can see that for $M=1$ (i.e., the STP model) the inferred trajectory is roughly an average of straight and right-turn modes. Incrementing the number of modes to $2$ we get a clear separation between modes going straight and turning right. Further, setting $M$ to $3$ the left-turn mode also appears, albeit with lower probability. When we set $M=4$ we get an interesting effect, where the go-straight mode splits into "fast" and "slow" modes, modeling variability in actor's longitudinal movement. We noticed the same effect on straight roads far from intersections, where the straight mode would split into several modes accounting for different velocities.

Lastly, we analyzed calibration of the predicted mode probabilities.
In particular, using test data we computed the relationship between predicted mode probability and likelihood that the mode is the best matching one to the ground-truth trajectory (i.e., that the mode is equal to $m^*$ as defined in equation \eqref{eq:best_m}).
We bucketed the trajectories according to their predicted probability and computed the average mode-matching likelihood for each bucket.
Figure \ref{fig:conf_calib} gives the result when we use $M=3$, while results for other values of $M$ resemble the ones shown.
We can see that the plot closely follows the $y=x$ reference line, indicating the predicted probabilities are well-calibrated.


\section{Conclusion}
Due to the inherent uncertainty of traffic behavior, autonomous vehicles need to consider multiple possible future trajectories of the surrounding actors in order to ensure a safe and efficient ride. In this work, we addressed this critical aspect of the self-driving problem and proposed a method to model the multimodality of vehicle movement prediction. The approach first generates a raster image encoding surrounding context of each vehicle actor and uses a CNN model to output several possible trajectories along with their probabilities. We discussed several multimodal models and compared to the current state-of-the-art methods, with the results strongly suggesting practical benefits of the proposed approach.
Following extensive offline evaluation, the method was successfully tested onboard SDVs in closed-course tests.




\bibliographystyle{IEEEtran}
\bibliography{icra_2019}

\end{document}